\documentclass[letterpaper]{article} 
\usepackage{aaai23}  
\usepackage{times}  
\usepackage{helvet}  
\usepackage{courier}  
\usepackage[hyphens]{url}  
\usepackage{graphicx} 
\usepackage{natbib}  
\usepackage{caption} 
\usepackage{algorithm}
\usepackage{algorithmic}
\usepackage{newfloat}
\usepackage{listings}
\DeclareCaptionStyle{ruled}{labelfont=normalfont,labelsep=colon,strut=off} 
\floatstyle{ruled}
\newfloat{listing}{tb}{lst}{}
\floatname{listing}{Listing}
\usepackage{times,latexsym}
\usepackage{url}
\usepackage[T1]{fontenc}

\usepackage{graphicx}
\usepackage{calrsfs}
\usepackage[T1]{fontenc}
\usepackage{multirow}
\usepackage[utf8]{inputenc}
\usepackage{microtype}
\usepackage{amsfonts}
\usepackage{inconsolata}

\DeclareMathAlphabet{\pazocal}{OMS}{zplm}{m}{n}

\newcommand{\standardentail}{\texttt{TE}}
\newcommand{\contextmodel}{\texttt{Context-TE}}
\newcommand{\parallelmodel}{\texttt{Parallel-TE}}

\title{Learning to Select from Multiple Options}
\author {
    Jiangshu Du\textsuperscript{\rm 1},
    Wenpeng Yin\textsuperscript{\rm 2},
    Congying Xia\textsuperscript{\rm 3},
    Philip S. Yu\textsuperscript{\rm 1}
}
\affiliations {
    \textsuperscript{\rm 1}University of Illinois at Chicago, Chicago, IL, USA \\
    \textsuperscript{
    \rm 2}Penn State University, State College, PA, USA \\
    \textsuperscript{
    \rm 3}Salesforce Research, Palo Alto, CA, USA\\
    
    \{jdu25, psyu\}@uic.edu, wenpeng@psu.edu, c.xia@salesforce.com
}
    
\begin{document}

\maketitle

\begin{abstract}
Many NLP tasks can be regarded as a selection 
problem from a set of options, e.g., classification tasks, multi-choice QA, etc. 
Textual entailment (\standardentail) has been shown as the state-of-the-art (SOTA) approach to dealing with those selection problems. 
\standardentail~treats input texts as premises (\texttt{P}),  options  as hypotheses (\texttt{H}), then handles the selection problem by modeling (\texttt{P}, \texttt{H}) pairwise. Two limitations: (i) the pairwise modeling  is unaware of other options, which is less intuitive since humans often  determine the best options by comparing  competing candidates; (ii) the inference process of pairwise \standardentail~is  time-consuming, especially when the option space is large. To deal with the two issues, this work  first proposes a contextualized  \standardentail~model (\contextmodel) by appending other $k$ options as the context of the current (\texttt{P}, \texttt{H}) modeling.  \contextmodel~is able to learn more reliable decision for the \texttt{H} since it considers various context. Second, we speed up \contextmodel~by coming up with \parallelmodel, which learns the decisions of multiple options simultaneously. \parallelmodel~significantly improves the inference speed while keeping comparable performance with \contextmodel. 
Our methods are evaluated on three tasks (ultra-fine entity typing, intent detection and multi-choice QA) that are typical selection problems with different sizes of options. Experiments show our models set new SOTA performance; particularly, \parallelmodel~is faster than the pairwise \standardentail~by $k$ times in inference.
Our code is publicly available at https://github.com/jiangshdd/LearningToSelect.
\end{abstract}

\section{Introduction}
\label{intro}

The NLP consists of  various  tasks and many of them are essentially a selection problem from a set of options. For instance, text classification selects the correct labels for the input text from all  label candidates; given a paragraph and multiple answer candidates, multi-choice question answering (QA)  selects the correct answer among all the choices.
Textual entailment (\standardentail) has been widely applied as the SOTA technique for solving these selection problems, e.g.,   intent detection \cite{fstc}, ultra-fine entity typing \cite{lite}, coreference resolution \cite{yinentailment},  relation extraction \cite{fstc,DBLPSainzLLBA21}, event argument extraction \cite{DBLP01376}, etc.

For those selection tasks, traditional classifiers treat all labels as indices, ignoring their semantics which, however, are the core supervision in zero- and few-shot learning. 
In contrast, \standardentail~reserves and exploits the original label information by constructing premise-hypothesis pairs, where premises (\texttt{P}) are texts and hypotheses (\texttt{H}) are transformed from labels.
Then, \standardentail~selects an option by predicting if an option-oriented \texttt{H} can be entailed by the input \texttt{P}. In addition to using a unified textual entailment framework to solve various selection problems, another benefit is that the availability of large-scale entailment datasets, such as MNLI~\cite{mnli} and DocNLI~\cite{docnli}, can provide rich indirect supervision  to  handle those target problems when task-specific supervision is limited. However, there are two main limitations for the standard \standardentail~methods.
First,  \standardentail~pairs one input text only with a single option. Thus, all options are treated independently when the model makes decisions. In contrast, humans usually perform the selection in a more intuitive way: comparing all options and select the best one. 
Second, at the inference phase, the  \standardentail~model needs to compare a text with all possible options one by one, which is inefficient especially when the option space is large. Take the  ultra-fine entity typing \cite{ufet} as an example, its  10k types  take the \standardentail~model 35 seconds for each test instance and about 19.4 hours~\footnote{Experiments run at an NVIDIA TITAN RTX.} to infer the entire test set~\cite{lite}.

To overcome the limitations of standard \standardentail, we introduce two novel approaches for option inference. First, inspired by the  intuition that a model can be more powerful if it can  find the correct answers even under the disturbance from other options, we propose a contextualized \standardentail~model (\contextmodel). It appends other $k$ options as an extra context of the standard (\texttt{P}, \texttt{H}) pairs during training. In this way, the model makes the decision not only depending on the current option $\texttt{H}$ but also considering a more informative context. Thus, the prediction on \texttt{H} is more reliable, but \contextmodel~is as slow as the standard \standardentail~in terms of inference.
Second, we improve the efficiency of \contextmodel~by  introducing a parallel \standardentail~method (\parallelmodel).
\parallelmodel~learns an option's representation based on other options and makes the decisions of $k$ options simultaneously. Therefore, the inference time of \parallelmodel~is faster than \contextmodel~by $k$ times.

We evaluate our proposed models on three tasks: ultra-fine entity typing (UFET \cite{ufet}, few-shot intent detection (BANKING77 \cite{banking77}, and multiple-choice QA (MCTest \cite{mctest}). 
The three tasks are representative selection problems: long texts and small-size options (MCTest), medium-size options (77 intents in BANKING77), and large-size options with multi-label selection (UFET has  over 10,000 types and multiple can be correct for a given entity mention). Our proposed \contextmodel~sets the new state-of-the-art performance on all three tasks, and the \parallelmodel~sacrifices a little bit performance (except for the UFET) while showing clear efficiency gains.

Our contributions can be summarized as the following three points. First, we discuss the limitations of the standard \standardentail~method in dealing with selection problems and propose two novel learning to select models---\contextmodel~and \parallelmodel---to overcome those issues.  Second, our experiments show that both \contextmodel~and \parallelmodel~outperform the standard \standardentail, and set the new state-of-the-art performance on multiple benchmarks. Third, we provide a deep analysis to better understand why the new models work. 

\section{Related Work}

Our work is mainly related with textual entailment and how it is applied to solve other NLP tasks.
\paragraph{Textual entailment.}  \citet{pascal} first introduced the concept of \standardentail~and released a challenging benchmark, \emph{Recognizing Textual Entailment} (RTE), for it. Then it attracted many following studies and the early stage of \standardentail~study  mainly focused on  lexical and syntactic level features \cite{DBLPAndroutsopoulosM10,DBLPReiB11}.
In recent years, many  large-scale \standardentail~datasets are released such as  SNLI~\cite{snli}, MNLI~\cite{mnli}, SciTail \cite{DBLPKhotSC18}, ANLI \cite{DBLPeWDBWK20} etc., which greatly advance the study of sentence-level \standardentail. \citet{docnli} also introduced a document-level dataset named DocNLI, which is constructed by reformatting and aggregating some NLP tasks.
The recent \standardentail~systems mainly rely on pairwise modeling over the premise and hypothesis using attentive recurrent neural networks \cite{DBLPRocktaschelGHKB15,DBLPWangJ16,DBLPWangHF17}, attentive convolutional neural networks \cite{DBLPSantosTXZ16,DBLPYinS18}, and pretrained transformers \cite{DBLPDevlinCLT19,DBLP11692}. Our work is related to \standardentail~but more interested in applying \standardentail~to solve downstream selection problems.

\paragraph{Textual entailment solves other NLP tasks.} Since many NLP tasks can be converted into a \standardentail~problem, \standardentail~naturally can provide indirect supervision for solving those tasks, especially when the task-specific supervision is limited. \citet{DBLPYinHR19} presented the first work that used \standardentail~supervision to solve zero-shot text classification. The idea was then applied to handle  few-shot intent identification \cite{dnnc,fstc}, ultra-fine entity typing \cite{lite}, coreference resolution \cite{yinentailment},  relation extraction \cite{fstc,DBLPSainzLLBA21}, event argument extraction \cite{DBLP01376}, zero-shot machine comprehension \cite{docnli}, etc. \citet{entailment_as_few_shot_learner} directly claimed that \standardentail~is a few-shot leaner for a wide range of NLP tasks.

All the literature we discussed above follow the standard \standardentail~method to solve  selection problems, i.e., inferring the options one by one.
This method suffers from two  limitations as we stated in Introduction.
In this work, we try to enhance the representation learning of 
\standardentail~and improve its inference speed in dealing with large option spaces.

\section{Methods}
\begin{figure*}
\centering
  \includegraphics[width=0.7\linewidth]{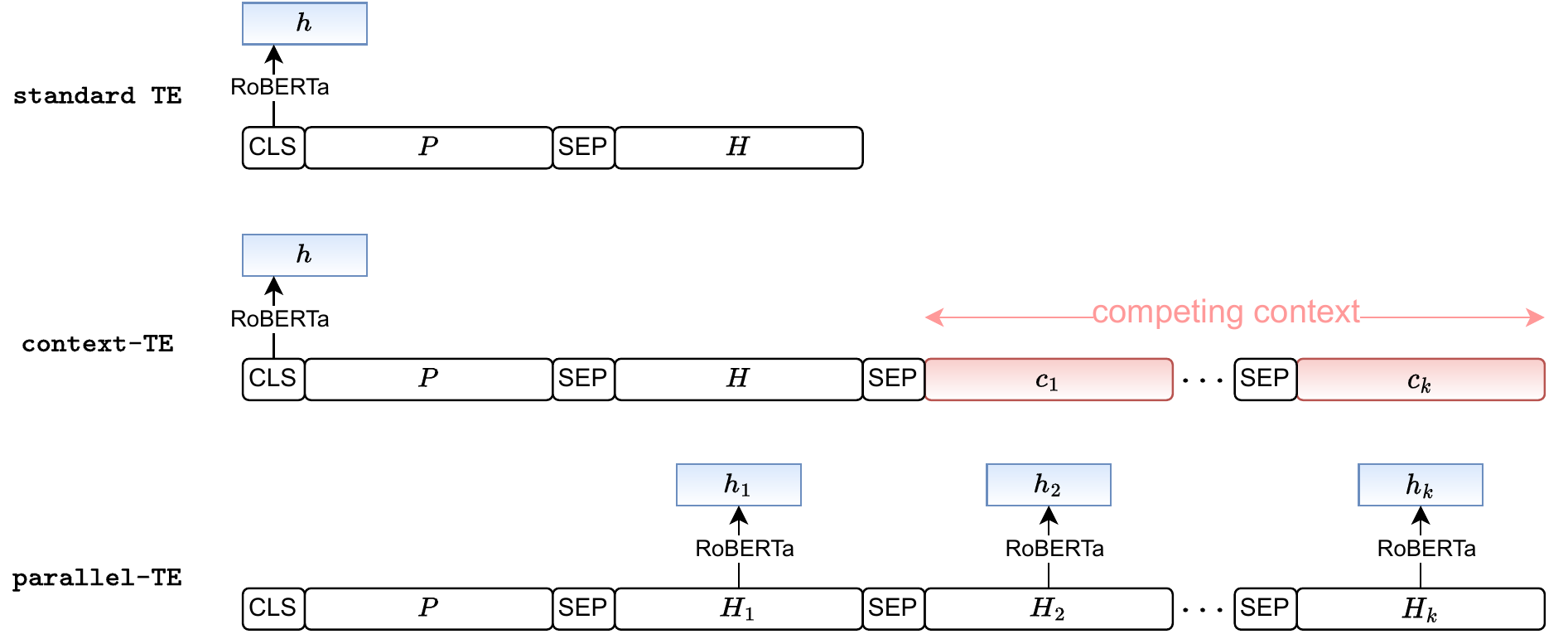}
  \caption{The comparison among \standardentail, \contextmodel~and \parallelmodel~in terms of their inputs and the representation ($h$) learning for hypotheses $H$.}
  \label{fig_model_comparison}
\end{figure*}

 This section presents our approaches to strengthening the options' representation learning (Section ``\contextmodel'') and speeding up the selection process among large size of options (Section ``\parallelmodel''). 

\subsection{Problem Definition}\label{sec:taskdefinition}

We formulate the selection problem as follows:
given a text $t$ and an option space $\mathbb{O}$ consisting of $n$ options $\{o_1, o_2, \cdots, o_{n}\}$, selecting the best  (single-label) or more (multi-label) options that match the $t$ under the definition of the task. Standard \standardentail~methods treat the $t$ as the premise (\texttt{P}) and an option as a hypothesis (\texttt{H}). For each $t$,  \standardentail~considers the option $o_i$ without comparing $o_i$ with other competing options. When the $n$ is large, each $t$ needs the pairwise modeling ($t$, $o_i$) totally $n$ times in inference. In Section ``\contextmodel'', we first present our method \contextmodel~to encode the interactions among options when modeling a particular option.   

\subsection{\contextmodel}
\label{sec:contextmodel}

\contextmodel~is a  contextualized representation learning paradigm for \standardentail. It adds other options as the  context when modeling the pair (\texttt{P}, \texttt{H}).

\textbullet \textbf{Contextualizing \standardentail~ pairs.} \contextmodel~appends $k$ options from $\mathbb{O}$ in a random order after  the original (\texttt{P}, \texttt{H}) as a \textit{competing context} ($c_1, c_2, \cdots, c_k$),  resulting in a contextualized pair: (\texttt{P}, \texttt{H}, $c_1$, $c_2$, $\cdots$, $c_k$). Note that the options resulting in $c_i$ ($i=1,\cdots,k$) should not be the same as the option generating \texttt{H}. The competing context introduces the interactions between  \texttt{H} and  other $k$ candidates in the option space in a single training instance so the model can better distinguish  similar options. The gold entailment label (i.e., \texttt{entailment/contradict/neutral}) of the contextualized pair (\texttt{P}, \texttt{H}, $c_1$, $c_2$, $\cdots$, $c_k$) is set the same as  (\texttt{P}, \texttt{H}). 
For instance, in an intent detection task, given an utterance text ``\textit{What's go on, where is my new card?}'' and its gold intent  ``\textit{card arrival}'', a standard \standardentail~pair (``\textit{What's go on, where is my new card?}'', ``\textit{card arrival}'') can be contextualized as (``\textit{What's go on, where is my new card?}'', ``\textit{card arrival}'', ``\textit{lost or stolen card}'', ``\textit{atm support}'') if $k=2$, and the relation between the option ``card arrival'' and the utterance ``\textit{What's go on, where is my new card?}'' should not change despite the existence of  competing context. The rationale of \contextmodel~is that if \texttt{P} can entail \texttt{H} in various contexts, \texttt{H} is more likely to be true given \texttt{P}.

\textbullet \textbf{Training.} As Figure \ref{fig_model_comparison} (middle) illustrates, we feed contextualized pairs into the encoder RoBERTa~\cite{DBLP11692}. The representation corresponding to the CLS token is used to denote the \texttt{H} in the context of \texttt{P} as well as the competing context ($c_1, c_2, \cdots, c_k$).  The remaining classification architecture and training process are the same with standard \standardentail~that works on (\texttt{P}, \texttt{H}).

\textbullet \textbf{Inference.} The inference of \contextmodel~is the same as the standard \standardentail. First, we convert the test set of the target task into (\texttt{P}, \texttt{H}) pairs by putting $t$ with each option $o_i$ together. Note that no competing context is needed for inference. Next, all the pairs are fed into the trained model and an entailment score for each pair will be given. For a single-label task, the option  with the highest score will be returned. For a multi-label task, all the options  with scores higher than a threshold $\tau$ are returned. Same as \standardentail, \contextmodel~also needs to model $x\times n$ sizes of (\texttt{P}, \texttt{H}) pairs if there are $x$ test inputs and $n$ options. This time-consuming process motivates our second method \parallelmodel.

\subsection{\parallelmodel}\label{sec:parallelmodel}
 
\parallelmodel~also works on the contextualized pairs (\texttt{P}, \texttt{H}, $c_1$, $c_2$, $\cdots$, $c_k$) except that all options, including \texttt{H} and $c_i$ ($i=1,\cdots,k$), are treated equally and optimized to learn their respective labels simultaneously.

\textbullet \textbf{Training.} Different from \contextmodel, all the options in the pair (\texttt{P}, \texttt{H}$_1$,  $\cdots$, \texttt{H}$_k$) will keep their original labels ($y_i$ for \texttt{H}$_i$) for joint training. Then the task is transformed to: given a sequence consisting of a premise and $k$ hypotheses,  select the correct hypotheses.

The input format of \parallelmodel~is the same as \contextmodel~but they have different representation learning processes. As shown in Figure \ref{fig_model_comparison} (bottom), we first feed (\texttt{P}, \texttt{H}$_1$,  $\cdots$, \texttt{H}$_k$)  into RoBERTa and obtain a series of token-level representations on the top layer. Then for each \texttt{H}$_i$ ($i=\{1, \cdots, k\}$), we average the representation vectors of all its tokens element-wise as its  representation $h_i$:
\begin{equation}
    h_i = \frac{1}{T}\sum_{j=1}^T \mathbf{RoBERTa}(\texttt{H}_i^j),
\end{equation}
where \texttt{H}$_i$ has $T$ tokens, and \texttt{H}$_i^j$ is the $j$-th one.

Next, each \texttt{H}$_i$ will learn a score $s_i$ ranging from 0 to 1, indicating how likely \texttt{H}$_i$ is entailed by \texttt{P}, by  feeding $h_i$ into a MLP:
\begin{equation}
    s_i = \mathrm{sigmoid}(\mathrm{MLP}(h_i))
\end{equation}

The loss $l$ of \parallelmodel~is  defined as the binary cross entropy (BCE) over all \texttt{H}$_i$ ($i=1,\cdots,k$):
\begin{equation}
    l = -\frac{1}{k}\sum_{i=1}^k y_i \cdot \mathrm{log} s_i + (1-y_i)\cdot \mathrm{log}(1-s_i)
\end{equation}

\textbullet \textbf{Inference.} For each hypothesis \texttt{H}$_i$ in the input, \parallelmodel~gives an entailment score. Then we can select the option of highest score for single-label tasks, or the options scored higher than a certain threshold for   multi-label tasks. Because \parallelmodel~models $k$ hypotheses simultaneously for a single \texttt{P}, its inference  is $k$ times faster than the standard \standardentail~and \contextmodel. More analyses are shown in Section ``Inference speed of \parallelmodel''.

\paragraph{\contextmodel~vs. \parallelmodel.} They have the same input structures as illustrated in Figure~\ref{fig_model_comparison}. Both of them learn to select the correct option upon comparing with other options. They mainly differ in two aspects: i) \contextmodel~only models one hypothesis in each contextualized pair while \parallelmodel~treats all the options in a pair as hypotheses, and infers all of them at the same time, which brings a huge boost regarding the inference efficiency; ii) \parallelmodel~adopts BCE loss to deal with multi-label selection tasks. In contrast, \contextmodel~uses the cross entropy loss as each pair only needs to be classified as entailment or non-entailment.

For the real-world problems where many options exist such as ultra-fine entity typing and open-world intent detection, it is not feasible to embed all the options into a single pair due to the input length limit of the encoder. Therefore, we first retrieve top-$k$ options to reduce the option space for those tasks. Details are discussed in Section `` Experiments''.

\section{Experiments}
\label{sec_experiments}

Our experiments are conducted on three different tasks:  \emph{ultra-fine entity typing}, \emph{few-shot intent detection}  and  \emph{multiple-choice QA}. 
We choose the three tasks since they represent different selection problems in NLP.
Ultra-fine entity typing is a multi-label task with a large option space: over 10,000 entity types. 
The few-shot intent detection task evaluates our proposed models under few-shot selection scene.
Multiple-choice QA is a selection problem which requires the model to understand long paragraphs. 

\paragraph{Top-$k$ options generation.} For the tasks with a large option space, it is infeasible to encode all options in the same input. We first  find the top-$k$ options for each text with an efficient method before the \contextmodel~and \parallelmodel~steps, hoping that the top-$k$ options are the most promising ones  for the input text. The kept $k$ options should have a high recall as we do not want the test inputs miss the gold options too much at this stage.

Concretely, for the ultra-fine entity typing task, we fine-tuned a bi-BERT \cite{DBLPDevlinCLT19} on the training data with one BERT encoding an entity mention and another encoding a type candidate. For the intent detection task, we directly use the Sentence-BERT \cite{sentence_bert} to match an utterance and an intent candidate because utterances and their gold intents usually have high sentence similarities. 
The top-$k$ model selection is based on the recalls on tasks' $dev$ set.
This step does not apply to the multi-choice QA task since there are only four answer candidates for each question.  

The reason we choose bi-BERT or pretrained Sentence-BERT is that these representation learners decouple the \texttt{P} and \texttt{H} and use separate encoders to model  them; as a result, they just need to generate representations for all options once and our model can reuse the same options' representations to compare with all inputs. After this step, each text $t$ has $k$ options that are most likely to be true.
The top-$k$ recall and the analysis about the influence of different $k$ values are reported in Section ``Influence of $k$ values''.

\paragraph{Model configurations.} We use the public pretrained  RoBERTa-large-mnli model as our backbone for the entity typing and intent detection tasks, and  the pretrained  RoBERTa-large on DocNLI~\cite{docnli} for the QA task for a fair comparison. The hyperparameters, threshold $\tau$, and $k$ are searched on the $dev$ set for each task.

\subsection{Ultra-Fine Entity Typing}
\label{entity_typing}

\paragraph{Dataset.} In this task, we use the ultra-fine entity typing (UFET) benchmark \cite{ufet}.  It has 5,994 human-annotated examples and 10,331 labels; each entity mention may have multiple types that are correct. The annotated examples are equally split into $train$, $dev$ and $test$. In addition, 
UFET  provides distant supervision data as an extra training resource, but we only train our models on the human-annotated dataset. The official evaluation metric is F1.

\paragraph{Baselines.} The following  prior systems are compared:

\begin{itemize}
    \item \textbf{LDET}~\cite{ldet} trains a learner to clean the distant supervision data. The learner discards the unusable data and fixes the noisy data. The denoised distant supervision data then is added to train a bi-LSTM \cite{DBLPHochreiterS97} incorporated with pretrained ELMo \cite{DBLPPetersNIGCLZ18} representations.
    \item \textbf{Box}~\cite{box} leverages BERT to project entity mentions and types as box embeddings, which can better capture the hierarchical relationships between fine-grained entity types. Predictions are finalized according to the embedding intersection in the box embedding space.
    \item \textbf{LRN}. \citet{lrn} proposed a label reasoning network, leveraging auto-regressive networks and bipartite attribute graph to recognize the extrinsic and intrinsic dependencies among entity types. By exploiting the dependency knowledge and reasoning, more correct labels can be discovered at the inference stage. 
    \item \textbf{MLMET}~\cite{mlmet} leverages the inner knowledge retained by a pretrained BERT model. The authors first used a BERT-based Masked Language Model (MLM) to generate extra distant supervision data and then trained a BERT model on three resources: the human-annotated, distant supervision, and MLM-generated data.
    \item \textbf{LITE}~\cite{lite} is the previous SOTA model. It converts the entity typing problem into a \standardentail~task  and uses a RoBERTa-large model pretrained on MNLI~\cite{mnli} as the backbone. LITE treats entity-mentioning sentences as premises. Type options are first transformed to statements based on the template ``[ENTITY] is a [LABEL]'' and then are treated as hypotheses. It also adds label dependency in \standardentail. A margin ranking loss is used to rank the entailment pairs over the non-entailment ones. \emph{All \standardentail-related models in this work, including \contextmodel~and \parallelmodel, use the same template as the baseline ``LITE''.}

\end{itemize}
\begin{table}[t]
\centering
\setlength{\tabcolsep}{5pt}
\begin{tabular}{l|ccc} 
\hline
Model & P & R & F1 \\ 
\hline
LDET~{\small\cite{ldet}} & 51.5 & 33.0 & 40.1 \\
Box~{\small\cite{box}} & 52.8 & 38.8 & 44.8 \\
LRN~{\small\cite{lrn}} & \textbf{54.5} & 38.9 & 45.4 \\
MLMET~{\small\cite{mlmet}} & 53.6 & 45.3 & 49.1 \\
LITE~{\small\cite{lite}} & 52.4 & 48.9 & 50.6 \\
\hline
\contextmodel & 53.7 & 49.4 & 51.5 \\
\parallelmodel & 54.0 & \textbf{51.0} & \textbf{52.4} \\
\hline
\end{tabular}
\caption{Results on the UFET task.}
\label{ufet_result}
\end{table}

\textbf{Results.} Table~\ref{ufet_result} shows the performance of our models and baselines. The $k$ value is selected on the $dev$ set and set to 80 for both \contextmodel~and \parallelmodel. 

First, we notice that both our approaches \contextmodel~and \parallelmodel~get new SOTA performance on this task with the \parallelmodel~performs the best (52.4). This is particular impressive if we highlight that those baselines make use of either extra   weak supervision data (e.g., LDET, MLMET) or the hierarchical relationships among types (e.g., Box, LRN, LITE). Second, even compared with LITE, the prior SOTA system that adopts the \standardentail~framework as well, \contextmodel~is able to learn better representations for hypotheses since it encoded the competing context.

\subsection{Intent Detection}
\label{intent_detection}

\paragraph{Dataset.} We use the BANKING77 dataset \cite{banking77} for the intent detection task. It is in the domain of online banking queries and consists of 77 intents. BANKING77 contains 10,003 training and 3,080 testing examples. We explore the few-shot learning ability of our models in this task. From the training data, we randomly sample 5-shot and 10-shot instances per intent as our $train$ respectively. We also sample a small portion of the training dataset as our $dev$, following the previous setting~\cite{cpft, dialoglue}. All experiments are run three times with distinctly sampled $train$ and we report the average performance, evaluated by accuracy.

\paragraph{Baselines.} We compare  with the following baselines:

\begin{itemize}
    \item \textbf{DualEncoder}~\cite{banking77} is a dual sentence encoder model combined with USE~\cite{use} and ConveRT~\cite{convert}. Both USE and ConverRT are strong sentence encoders pretrained on different conversational response tasks. The combination of both yields better performance than each of them.
    \item \textbf{ConvBERT+}~\cite{convbert_combined} is based on  ConvBERT~\cite{dialoglue}, a BERT model pretrained on a large dialogue corpus. By combining with example-driven training, task-adaptive training and observers, the model obtains a strong performance.
    \item \textbf{DNNC}~\cite{dnnc} is a discriminative nearest-neighbor model pretrained on three different \standardentail~datasets: SNLI~\cite{snli}, MNLI~\cite{mnli}, and WNLI~\cite{wnli}.
    \item \textbf{CPFT}~\cite{cpft} adopts  RoBERTa as the backbone and first conducts self-supervised contrastive pretraining on six intent detection datasets. The model is then fine-tuned on few-shot data with supervised contrastive learning. \emph{For a fair comparison, we report their performance without the contrastive pretraining on extra data.}
    \item \textbf{\standardentail}~\cite{fstc} converts intent detection to traditional \standardentail~by treating utterances and intent options as premises and hypotheses, respectively.

\end{itemize}

\begin{table}[t]
\setlength{\tabcolsep}{2pt}
\centering
\begin{tabular}{l|cc} 
\hline
Model & 5-shot & 10-shot \\ 
\hline

CPFT \cite{cpft} & 76.75 & 84.83 \\
DualEnc. \cite{banking77}  & 77.75 & 85.19 \\
ConvB. \cite{convbert_combined} & - & 85.95 \\
DNNC \cite{dnnc} & 80.40 & \textbf{86.71} \\ 

\standardentail~\cite{fstc} & 78.21 & 82.51 \\ 
\hline
\contextmodel & \textbf{80.76} & 85.53 \\
\parallelmodel & 78.69 & 83.29 \\
\hline
\end{tabular}
\caption{Few-shot performance on BANKING77. }
\label{banking77_results}
\end{table}
\paragraph{Model details.} Since the intent detection task is under the few-shot setting, data augmentation can be helpful during the training stage. \contextmodel~is naturally suitable for augmenting data because it expands the training data by introducing the extra (\texttt{P}, \texttt{H}) pairs equipped with competing context. 
For \parallelmodel, we perform data augmentation as follows: for each pair with $k$ hypotheses, we shuffle it $k$ times, making sure the positive option is at the different positions in all new sequences. This is to let the model learn an option's gold label wherever the option is located in the input sequence.

In this task, \contextmodel~infers at the full label space $\mathbb{O}$ ($|\mathbb{O}| = 77$). Due to the max sequence length limitation of RoBERTa, \parallelmodel~infers only the top-$k$ label options where $k$ ranges from 10 to 60, selected on $dev$. 

\paragraph{Results.} Test accuracy under 5-shot and 10-shot settings on BANKING77 benchmark is reported in Table~\ref{banking77_results}. 
\contextmodel~outperforms all the models  under the 5-shot setting and obtains competitive performance with the prior SOTA on 10-shot (85.53 vs. 86.71). \parallelmodel~demonstrates tiny drop of performance on both 5-shot and 10-shot. But compared with the standard \standardentail,  both \parallelmodel~ and \contextmodel~yield better results, suggesting that our proposed models are more effective than the traditional \standardentail.

\subsection{Multi-Choice Question Answering}
\label{qa}

\paragraph{Dataset.} We work on MCTest \cite{mctest},  a multiple-choice machine comprehension benchmark from a fictional story domain. One correct answer must be selected among four candidates given a question and a paragraph stating the background knowledge. Besides,  \citet{mctest} released a \standardentail~version  data by treating paragraphs as premises and converting  question-answer pairs into   hypotheses. 
MCTest is a low-resource task, consisting of two sets with 160 (MC160) and 500 (MC500) examples, respectively.
Our experiments are conducted on the \standardentail~version of MCTest. The official evaluation metric is accuracy.

\paragraph{Baselines.} We compare with three latest methods:

\begin{itemize} 
    \item \textbf{RDEC} \cite{DBLPYuZY19} designs an inferential network trained with reinforcement learning to understand contexts. This method consists of multiple attention-based reasoning steps and recursively constructs the evidence chain to improve the reasoning skill of machines.
    \item \textbf{\standardentail$_{\mathrm{DocNLI}}$}~\cite{docnli} is the previous SOTA system that first pretrains RoBERTa-large on DocNLI, a large corpus for document-level \standardentail, then finetunes on MCTest.
    \item \textbf{\standardentail$_{\mathrm{vanilla}}$} directly trains a RoBERTa-large on the \standardentail~version data without  pretraining on any extra \standardentail~datasets such as MNLI, DocNLI.
\end{itemize}

\paragraph{Model details.} In this task, \parallelmodel~ also performs data augmentation with the strategy elaborated in Section ``Intent detection'' because the training data is scarce. In addition,  \contextmodel~follows \standardentail$_{\mathrm{DocNLI}}$ to exploit the pretrained  RoBERTa on DocNLI as the encoder.
Note that this task does not require the top-$k$ option generation since the option size for each piece of data is merely four, but we allow the system to truncate the end of premises if the input length exceeds the max length limit of RoBERTa.

\begin{table}
\centering
\begin{tabular}{l|cc} 
\hline
Model & MC160 & MC500 \\ 
\hline
\standardentail$_{\mathrm{vanilla}}$ & 42.50 & 63.67 \\
RDEC \cite{DBLPYuZY19} & 80.00 & 75.50\\

\standardentail$_{\mathrm{DocNLI}}$ (Yin et al. 2021) & 90.83 & 90.66 \\ 
\hline
\contextmodel~& \textbf{92.50} & \textbf{91.67} \\
\parallelmodel~ & 81.25 & 82.50 \\
\hline
\end{tabular}
\caption{Test accuracy on MCTest.}
\label{mctest_results}
\end{table}

\paragraph{Results.} Table~\ref{mctest_results} shows the experiment results on MCTest.  Both \contextmodel~and \standardentail$_{\mathrm{DocNLI}}$ rely on the indirect supervision from DocNLI, but \contextmodel~outperforms \standardentail$_{\mathrm{DocNLI}}$ on both MC160 and MC500, achieving the new SOTA performance. \parallelmodel~performs worse than \contextmodel~because it is more comparable with the \standardentail$_{\mathrm{vanilla}}$:  both use the RoBERTa-large encoder and do not pretrain on extra data. Nevertheless, \parallelmodel~achieves multi-option joint training, which leads to the improvement by  large margins: 81.25 vs. 42.50 on MC160, and 82.50 vs. 63.67 on MC500.

\section{Analysis}

\subsection{What Contributes to the Improvement?}
\label{sec_ablation}
\begin{table}
\centering
\setlength{\tabcolsep}{1.5pt}
\begin{tabular}{c|ccc|cccc} 
\hline
\multirow{2}{*}{Model} & \multicolumn{3}{c|}{UFET} & \multicolumn{4}{c}{~ BANKING77} \\
 & P & R & F1 & 1-shot & 3-shot & 5-shot & 10-shot \\ 
\hline
\multicolumn{1}{l|}{\standardentail} & - & - & - & 67.53 & 73.74 & 78.21 & 82.51 \\
\multicolumn{1}{l|}{\standardentail$_{\mathrm{top}-k}$} & \textbf{54.4} & 47.7 & 50.8 & 63.95 & 72.54 & 77.31 & 80.54 \\ 
\multicolumn{1}{l|}{LITE$_{\mathrm{top}-k}$} & 49.6 & 51.0 & 50.3 & - & - & - & - \\
\hline
\contextmodel~& 53.7 & 49.4 & 51.5 & \textbf{69.40} & \textbf{77.42} & \textbf{80.76} & \textbf{85.53} \\
\parallelmodel~ & 54.0 & \textbf{51.0} & \textbf{52.4} & 69.11 & 75.12 & 78.69 & 83.29 \\
\hline
\end{tabular}
\caption{Ablation study on both ultra-fine entity typing and intent detection tasks. [MODEL]$_{\mathrm{top}-k}$ refers to the model runs on the retrieved top-$k$ options. \standardentail~runs on the full option space.}
\label{ablation}
\end{table}

The experimental results on the three benchmarks demonstrate the effectiveness of our proposed methods. However, it is still unclear: \textit{does the system benefit from the top-$k$ option generation as it reduces the option space or other inference strategies?} We conduct ablation study on both ultra-fine entity typing and intent detection tasks since our models need to perform the top-$k$ option generation  for them.

Specifically, given the top-$k$ selected options, we run other competitive systems  to see if the smaller option space helps them. For the entity typing task, we run the prior SOTA model, LITE,  on the same top-$k$ ($k=80$) as our  \parallelmodel~model. For the intent detection task, we run the standard \standardentail~method on the same top-$k$ ($k = 25$). Note that the \contextmodel~model infers in the entire option space on BANKING77. To gain a better insight into the impact of the top-$k$ generation step, we also extend our experiments with 1-shot and 3-shot settings. The data sampling and evaluation strategies do not change.

The ablation experiments results are shown in Table~\ref{ablation}. Given top-$k$ options,  both our systems \contextmodel~and \parallelmodel~turn to be the \standardentail$_{\mathrm{top}-k}$ if we discard the competing context and the multi-option joint training. However, we notice that \standardentail$_{\mathrm{top}-k}$ gets pretty close performance with the LITE$_{\mathrm{top}-k}$ on UFET (50.8 vs. 50.3 by F1) and even slightly worse performance than the standard \standardentail~on all the four few-shot settings of BANKING77. In contract, given the same top-$k$ options, both \contextmodel~and \parallelmodel~surpass the competitors with large margins. This means that our systems mainly benefit from the newly designed representation learning and training strategy rather than the smaller option space by top-$k$ generation.

\subsection{Influence of $k$ Values}
\label{k_analysis}

\begin{figure}[t]
  \centering
  \includegraphics[width=0.9\linewidth]{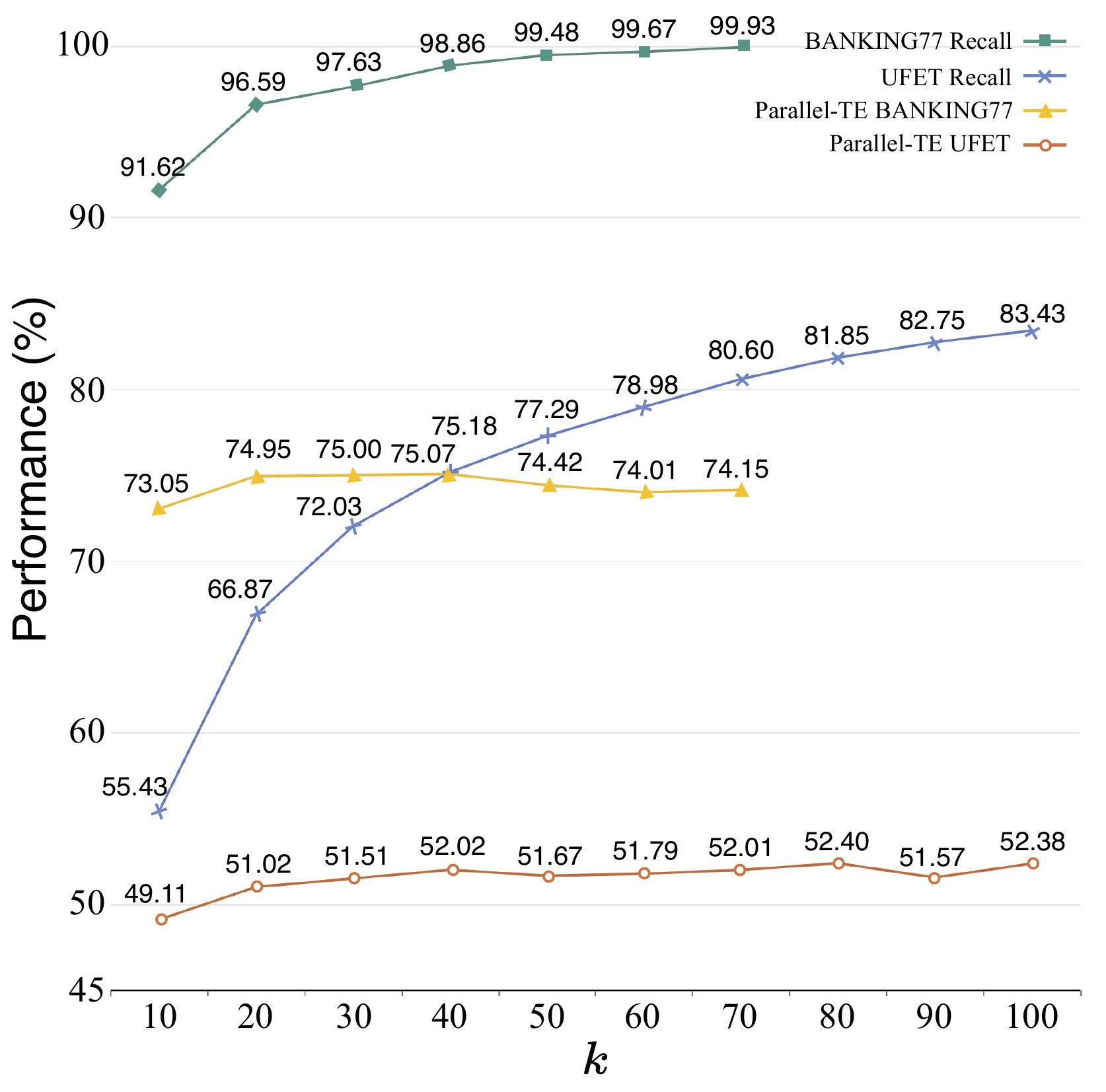}
  \caption{Top-$k$ recall and \parallelmodel~ performance on UFET and BANKING77 tasks when $k$ value varies. For the \parallelmodel~performance, we report F1 and accuracy on UFET and BANKING77, respectively.} 
  \label{fig_k_impact}
\end{figure}

A higher $k$ value can bring a higher top-$k$ recall; however, it also caps the model performance. In this section, we investigate how the $k$ values affects the \parallelmodel~performance.

As shown in Figure~\ref{fig_k_impact}, we report top-$k$ recall, where $k$ varies from 10 to 100 for UFET  and 10 to 70 for (3-shot) BANKING77  due to the max  length limit of RoBERTa. We also show the \parallelmodel~performance under different $k$ values. 
The top-$k$ recall on BANKING77 is already high (91.62\%) even when $k=10$, and its increasing speed slows down when $k>30$. \parallelmodel~ achieves the best performance with $k = 25$ as reported in the previous section. 
After that, a higher $k$ harms the \parallelmodel~performance. 
UFET top-$k$ recall gets more benefits from the increase of $k$. The highest F1 is achieved when $k=80$, and a very close performance is obtained when $k=100$.
These observations demonstrate a trade-off between $k$ values and the model performance. As $k$ increases, more correct options are captured in the top-$k$ option space, but it also brings more difficulties for \parallelmodel~to select the correct one since more competing options exist. Nonetheless, it is still beneficial to provide opportunities for those competing options to interact with each other so that the model can make better decisions.

\subsection{Influence of Top-$k$ Orders}

We study this factor in both training and inference.
\paragraph{Training.} As per our experiments, the order of top-$k$ options in a input instance is essential to \parallelmodel~at the training phase. We first start our experiments on UFET and keep the original order of the top-$k$ options. 
After top-$k$ generation, most correct options are at the front part since they usually receive higher scores.
Training a model on the pairs with the unshuffled top-$k$ options leads to a serious overfitting on the position information.
Thus, shuffling the order of top-$k$ is important for \parallelmodel~during training.
\paragraph{Inference.} We also investigate the influence of different top-$k$ orders at the inference stage by conducting experiments on few-shot BANKING77. 
Given the same text, different top-$k$ orders sometimes yield  predictions of tiny differences but mostly the predictions are consistent. This indicates that our training strategy in \parallelmodel~can result in very robust model behavior. Duplicating a pair with the shuffled top-$k$ orders and then making the final decision by majority voting improves the performance slightly, as shown in Table \ref{banking77_vote}, but these negligible improvements are not worth the time cost.

\begin{table}
\centering

\small
\begin{tabular}{l|cccc} 
\hline
Model & 1-shot & 3-shot & 5-shot & 10-shot \\ 
\hline
\parallelmodel& 69.11 & 75.12 & 78.69 & 83.29 \\
\enspace w/ majority vote & \textbf{69.89} & \textbf{75.70} & \textbf{79.31} & \textbf{83.57} \\
\hline
\end{tabular}
\caption{The results of \parallelmodel~and the majority voting on BANKING77.}
\label{banking77_vote}
\end{table}
\begin{figure}[t]
\centering
  \includegraphics[width=0.8\linewidth]{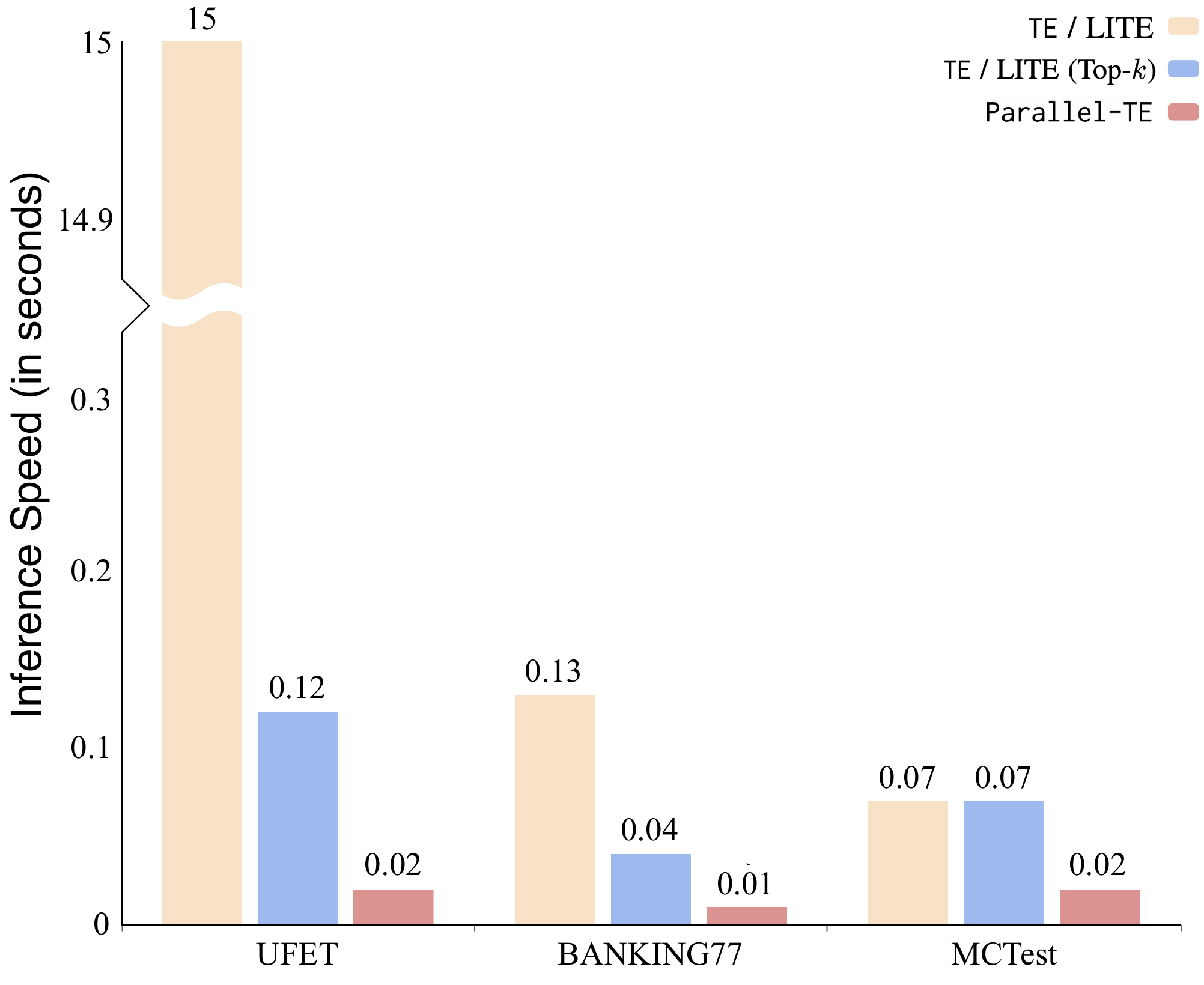}
  \caption{Inference speed (seconds/test case) of different \standardentail~models. \standardentail~/ LITE (Top-$k$) indicates that the model runs on the top-$k$ option space.}
  \label{fig_speed}
\end{figure}

\subsection{Inference Speed of \parallelmodel}
\label{sec_inference_speed}
We compare the inference speeds of \parallelmodel~with other entailment-based methods (\standardentail, LITE and their respective versions working on top-$k$ options) 
in Figure \ref{fig_speed}. \footnote{The inference speed is measured on an NVIDIA GeForce RTX 3090 with the evaluation batch size of 256.} 

Overall, the inference time of \standardentail/LITE over the three tasks decreases in the order of ``UFET $\rightarrow$ BANKING77 $\rightarrow$ MCTest'' since both methods search the gold option from the whole space and each subsequent task has a smaller size of options. The top-$k$ options can speed up \standardentail/LITE (i.e., \standardentail/LITE (top-$k$)) dramatically, which is within expectation. However, as we discussed in Section ``What contributes to the improvement?'', this operation may degrade the model performance. 

When apply our model \parallelmodel~on the top-$k$ options, the inference speed gets further boosted: on the UFET task, rerunning the prior SOTA model LITE takes 15 seconds to predict a single test instance, while our   model \parallelmodel~only spends 0.02 seconds. 

\subsection{Why \contextmodel~Works?}

\begin{figure}[t]
\centering
  \includegraphics[width=0.95\linewidth]{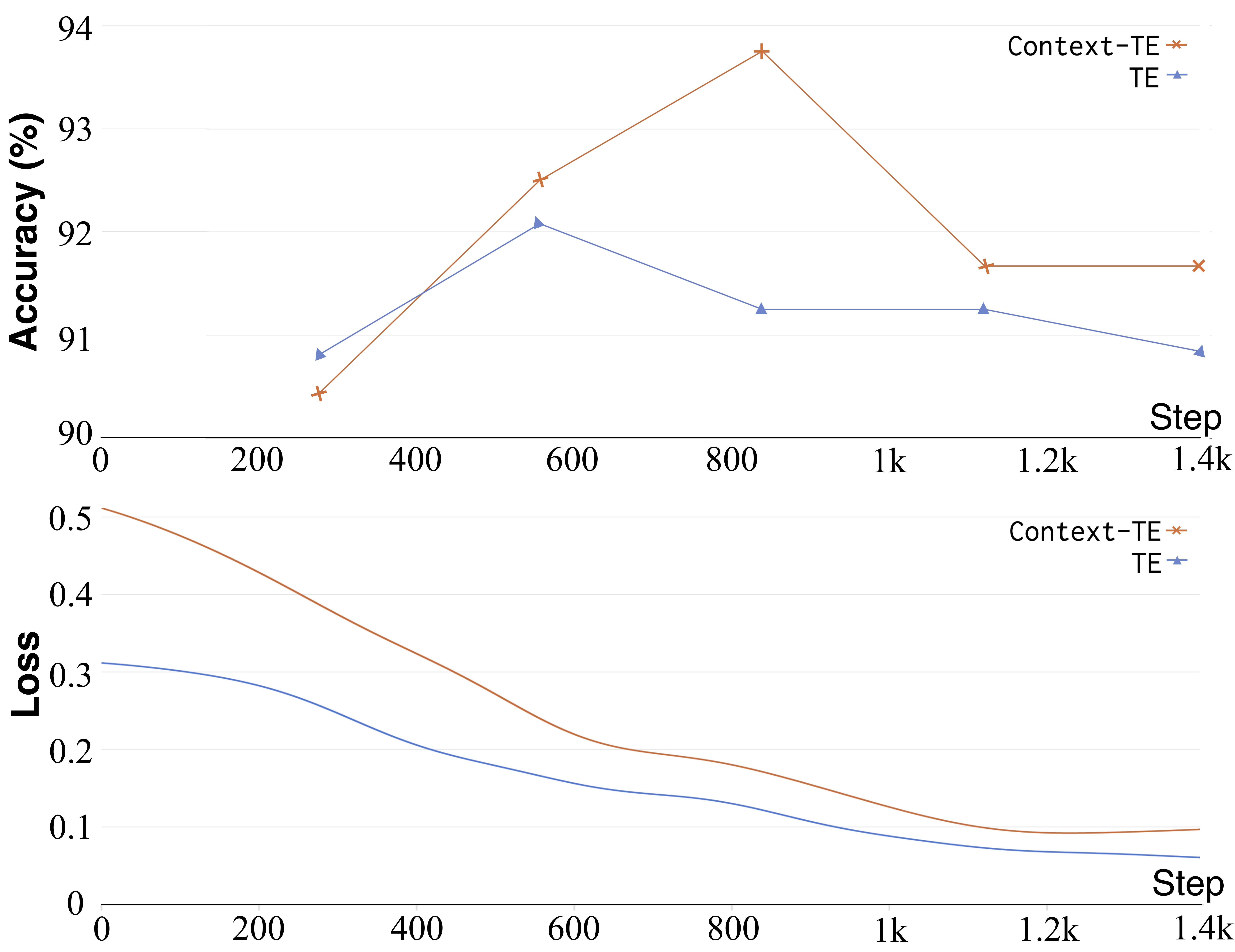}
  \caption{Training loss curves and test accuracy of \contextmodel~and \standardentail~on MC160 task. Loss curves are processed with Gaussian smoothing.}
  \label{fig_mc160_curve}
\end{figure}
By equipping (\texttt{P}, \texttt{H}) pairs  with competing context, \contextmodel~outperforms the standard \standardentail~model on all the three tasks.
We try to explain it by analyzing the training loss curves and test accuracy of both models on MC160 task, as shown in Figure~\ref{fig_mc160_curve}.
We train both models 5 epochs and report the test accuracy at the end of each epoch. The training loss is recorded at every training step and processed with Gaussian smoothing. 
As the figure illustrates, \contextmodel~always outperforms \standardentail~after the second epoch while holding  a higher training loss. This is reasonable because \contextmodel~introduces more challenging data (selecting the correct option under the disturbance of other options is harder), which acts as a regularizer that mitigates the 
overfitting  to some extent.

\section{Conclusion}
This work studied the issues of the popular \standardentail~framework in solving selection tasks (i.e., \emph{neglecting option-to-option comparison in representation learning} and \emph{low inference speed}) and proposed \contextmodel~and \parallelmodel~to solve them, respectively. Both new models outperform the standard \standardentail~method and mostly set the new state of the art performance in three typical tasks: ultra-fine entity typing, intent detection and multi-choice machine comprehension. 

\section{Acknowledgments}
The authors appreciate the reviewers for their insightful comments and suggestions.
This work is supported in part by NSF under grants III-1763325, III-1909323,  III-2106758, and SaTC-1930941. 
\bibliography{aaai23}

\end{document}